\documentclass[conference]{IEEEtran}
\IEEEoverridecommandlockouts

\usepackage{cite}
\usepackage{amsmath,amssymb,amsfonts}
\usepackage{algorithmic}
\usepackage{graphicx}
\usepackage{textcomp}

\usepackage{pifont}
\newcommand{\cmark}{\ding{51}}
\newcommand{\xmark}{\ding{55}}
\usepackage{array}
\usepackage[hidelinks]{hyperref}

\def\BibTeX{{\rm B\kern-.05em{\sc i\kern-.025em b}\kern-.08em
    T\kern-.1667em\lower.7ex\hbox{E}\kern-.125emX}}
    
\begin{document}

\title{Data Collection of Real-Life Knowledge Work\\ in Context: The RLKWiC Dataset}

\makeatletter
\newcommand{\linebreakand}{%
  \end{@IEEEauthorhalign}
  \hfill\mbox{}\par
  \mbox{}\hfill\begin{@IEEEauthorhalign}
}
\makeatother

\author{\IEEEauthorblockN{1\textsuperscript{st} Mahta Bakhshizadeh}
\IEEEauthorblockA{\textit{Department of Smart Data and Knowledge Services} \\
\textit{German Research Center for Artificial Intelligence (DFKI)}\\
Kaiserslautern, Germany\\
mahta.bakhshizadeh@dfki.de}\\
\linebreakand 
\IEEEauthorblockN{2\textsuperscript{nd} Christian Jilek}
\IEEEauthorblockA{\textit{Department of Smart Data and Knowledge Services} \\
\textit{German Research Center for Artificial Intelligence (DFKI)}\\
Kaiserslautern, Germany\\
christian.jilek@dfki.de}
\and
\IEEEauthorblockN{3\textsuperscript{rd} Markus Schröder}
\IEEEauthorblockA{\textit{Department of Smart Data and Knowledge Services} \\
\textit{German Research Center for Artificial Intelligence (DFKI)}\\
Kaiserslautern, Germany\\
markus.schroeder@dfki.de}
\and
\IEEEauthorblockN{4\textsuperscript{th} Heiko Maus}
\IEEEauthorblockA{\textit{Department of Smart Data and Knowledge Services} \\
\textit{German Research Center for Artificial Intelligence (DFKI)}\\
Kaiserslautern, Germany\\
heiko.maus@dfki.de}
\and
\IEEEauthorblockN{5\textsuperscript{th} Andreas Dengel}
\IEEEauthorblockA{\textit{Department of Smart Data and Knowledge Services} \\
\textit{German Research Center for Artificial Intelligence (DFKI)}\\
Kaiserslautern, Germany\\
andreas.dengel@dfki.de}
}

\maketitle

\begin{abstract}

Over the years, various approaches have been employed to enhance the productivity of knowledge workers, from addressing psychological well-being to the development of personal knowledge assistants. A significant challenge in this research area has been the absence of a comprehensive, publicly accessible dataset that mirrors real-world knowledge work. Although a handful of datasets exist, many are restricted in access or lack vital information dimensions, complicating meaningful comparison and benchmarking in the domain. This paper presents RLKWiC, a novel dataset of Real-Life Knowledge Work in Context, derived from monitoring the computer interactions of eight participants over a span of two months. As the first publicly available dataset offering a wealth of essential information dimensions (such as explicated contexts, textual contents, and semantics), RLKWiC seeks to address the research gap in the personal information management domain, providing valuable insights for modeling user behavior.

\end{abstract}

\begin{IEEEkeywords}
RLKWiC Dataset, Context-based user modeling, Knowledge work support, Personal information management.
\end{IEEEkeywords}

\section{Introduction}

Knowledge workers (KW) are individuals primarily concerned with processing and applying information rather than manual labor\cite{drucker1999knowledge}. Their main capital is knowledge\cite{davenport2005thinking} and they are often perceived as human subjects whose cognitive dimension is targeted with knowledge management systems and can be considered as investors of knowledge and energy in an organization \cite{reinhardt2011knowledge}.

Over the last few decades, researchers have striven to support KW in various capacities to enhance their productivity. While an ongoing research lane investigated supporting KW from a more psychological point of view (like enhancing well-being at work through stress-management \cite{soto2021observing, koldijk2016context}), other researchers have been focusing on KW's information needs towards developing Personal Knowledge Assistants (PKA). For instance, the idea of context-sensitive Proactive Information Delivery (PID) via a task-oriented view of an office worker’s personal knowledge space during daily knowledge-intensive tasks was proposed by Holz et al.\cite{HolzMausBernardiRostanin05b}. Although researchers have used different wordings, the idea of PID has been continued ever since. While some researchers have investigated information retrieval methods\cite{vuong2017watching, sappelli2017evaluation}, others have interpreted the idea as a Recommender System (RS) use case\cite{vuong2022behavioral, jacucci2021entity}.

Following this research lane, the concept of Contextual States (CS) has been recently introduced to leverage context-aware RS for improving PKA\cite{bakhshizadeh2021leveraging}. CS refer to multi-dimensional sessions in which all required contextual information needed to identify user's needs is taken into account. These dimensions consist of role, action, topic, task, files and apps, interaction behaviour, and semantics. The study argues that it's vital to observe users in various situations through these seven dimensions in order to provide KW with effective recommendations. Such recommendation methods can be evaluated either by investigating productivity improvements within an online experiment like implementing the EntityBot by Jacucci et al.\cite{jacucci2021entity} or using a preexisting offline knowledge work dataset for simulating such RS, which is the adopted approach by Sappelli, Verberne, and Kraaij \cite{sappelli2017evaluation}. Although the creation of a testbed for PIM-related experiments is a notably challenging task as it is highly complicated by privacy concerns \cite{chernov2007converting}, such a dataset can serve researchers not only as a benchmark for offline evaluation of PIM methods but also as a requirement for unlocking the vast potential of machine learning for PIM-related tasks\cite{dhar2013data}. The need for such a dataset in the domain has been addressed years ago by Chernov, Minack, and Serdyukov along with suggesting converting desktops into a personal activity dataset as a practical solution\cite{chernov2007converting}, which is the adopted approach in RLKWiC.

Although there are a few existing datasets in the domain, they often come with serious limitations. Some are not publicly accessible, while others lack crucial information dimensions needed for repeating the conducted experiments or extracting CS. The RLKWiC dataset on the other hand, is a step towards addressing this gap by being the first publicly available dataset setting a new benchmark for research in this domain. As RLKWiC encapsulates many of the vital information dimensions such as context, textual content accessed by participants (e.g. title and body of the browsed web pages), semantics, and knowledge actions, it can be beneficial in the mentioned research lanes towards improving PIM and supporting KW with their daily tasks.

In the following Section (\ref{RW}), the aforementioned existing datasets will be discussed and compared, highlighting the research gap that RLKWiC seeks to fill. The procedure of the conducted data collection for RLKWiC as well as the experiment design and tools will be detailed in Section \ref{DC}, followed by illustrating the structure and statistical overview of RLKWiC in Section \ref{DS}. Finally, the paper is concluded in Section \ref{CFW} along with briefly discussing the prospects for future works.

\section{Related Work} \label{RW}

A few datasets were collected for modeling KW' behaviors for different objectives during the last two decades as PKA and PIM services gained more attention. One aspect of supporting KW with their digital tasks in practice is to deal with the existing challenges of a real-life environment. These challenges include dealing with diverse topics, a high frequency of context switching\cite{gonzalez2004constant}, and a plethora of irrelevant injected data emerging from the user's distraction from the main topic. In addition, users show different behavior patterns which can be captured only in long-term observations\cite{bakhshizadeh2021leveraging}. Datasets that are collected in either a short time or in an experimental setting mostly fail to represent such challenges. While some of the existing datasets are collected in a controlled lab setting with pre-defined tasks \cite{koldijk2014swell, sappelli2014collecting, mirza2015study, satterfield2020identifying}, others rely on real-life data collected by installing monitoring tools that track participants’ computer interactions during their everyday digital activity \cite{sanchez2020behacom, pellegrin2021task, meyer2020detecting}. The observation periods range from a few hours per participant \cite{koldijk2014swell, sappelli2014collecting, mirza2015study, satterfield2020identifying, meyer2020detecting} to around two months \cite{sanchez2020behacom, pellegrin2021task}.

None of the aforementioned datasets (except for Meyer et al.\cite{meyer2020detecting} which provides some semantic information), feature any semantics. RLKWiC on the other hand, represents the user's information space through a personal Knowledge Graph (KG) based on the concept of the Personal Information Model (PIMO). PIMO is introduced by Sauermann et al.\cite{sauermann2007pimo} and has been developed within the last two decades of research towards self-organizing PKA \cite{jilek2023towards}. It is a framework of multiple ontologies to represent concepts and documents that are in the attention of the user when doing knowledge work. Basic concepts such as time, place, people, organizations, and tasks can be extended by the users at will to express their mental model\cite{sauermann2007pimo}.

The only dataset that features context is the SWELL knowledge work dataset of information behavior in context\cite{sappelli2014collecting} for stress and user modeling research\cite{koldijk2014swell}. What is considered as "contexts" in the SWELL dataset is in fact a set of pre-defined topics (such as "Stress at Work", "The Life of Napoleon", etc.) about which the participants were asked to write reports and prepare presentations. While RLKWiC adopts a more advanced definition of context introduced by Jilek et al.\cite{jilek2018context}.

Although SWELL provides many valuable features like contexts and written documents, it lacks some of the crucial features such as the content of the browsed web pages and semantics. Additionally, in contrast to RLKWiC, participants' activities are limited to eight predefined tasks in a short-term experimental setting. Furthermore, none of these datasets are publicly available except SWELL\cite{koldijk2014swell}, BEHACOM\cite{sanchez2020behacom}, and Pellegrin et al.'s dataset\cite{pellegrin2021task} which has published an excerpt of sample data with the full dataset on reasonable request.

The introduced datasets are summarized and compared based on the outlined features in Table \ref{comparison table}. As the table illustrates, these datasets published since 2014, were gathered either from real-life knowledge work or predefined tasks in a controlled experimental setting for diverse research objectives. Participants, numbering between five and 25, were primarily students or software developers. Data collection per participant spanned from two hours to two months, and the datasets exhibit different properties based on their collection intent. Most of the previously collected datasets do not provide contexts, textual contents, and semantics and are not completely open to the public. In contrast, RLKWiC is the first and currently only publicly available knowledge work dataset that features explicated user contexts, contains additional semantics and has all accessed information items (websites, files, etc.) available as part of the dataset.

\bgroup
\def\arraystretch{1.5}
\begin{table*}[ht]
\caption{Comparison of the existing KW datasets}
\begin{center}
\begin{tabular} 
{|p{1.1cm}|p{1.3cm}|p{0.5cm}|p{2cm}|p{1.7cm}|p{1.3cm}|p{2.1cm}|p{0.8cm}|p{1.0cm}|p{0.5cm}|p{1cm}|}
\hline
\textbf{}&\multicolumn{10}{|c|}{\textbf{Features}} \\
\cline{2-11} 

\textbf{Dataset}&
\textbf{\textit{Objective}}&
\textbf{\textit{Year}}&
\textbf{\textit{Settings/tasks}}&
\textbf{\textit{Participants}}&
\textbf{\textit{Amount}}&
\textbf{\textit{Properties}}&
\textbf{\textit{Context}}&
\textbf{\textit{Content}}&
\textbf{\textit{Sem}}&
\textbf{\textit{Available}} \\
\hline

SWELL\cite{koldijk2014swell, sappelli2014collecting} & Personalized support for well-being at work & 2014 & Experimental settings - pre-defined tasks including writing reports, making presentations, and reading emails & 25 Delf students and TNO interns, 8 females, 17 males, avg age 25, most native Dutch & About 3 hours per participant (about 75 hours in total) & Basic interaction events (e.g., mouse clicks, keystrokes, app switches), browsed webs URLs, and written documents & \cmark (set of pre- defined topics) & Written docs are available but web page contents are not & \xmark & \cmark\\
\hline

Mirza et al. \cite{mirza2015study} & Automatic classification of users’ desktop interactions & 2015 & Defined set of 20 activities from four typical activity models & 5 masters and PhD students in a computer science lab (avg age 27.5) & 6 hours each over one week & Operation type, timestamp, window handle and caption, app name, file path & \xmark & \xmark & \xmark & \xmark\\
\hline

Satterfield et al.\cite{satterfield2020identifying} & Identifying and describing information seeking tasks & 2020 & Controlled lab setting, 6 information seeking development Tasks & 17 software developers & 2 hours each (34 hours in total) & Screenshots of the active window at 1-second intervals. app names and window titles whenever changed & \xmark & \xmark & \xmark & \xmark\\
\hline

BEHA- COM\cite{sanchez2020behacom} & Modelling users’ behavior & 2020 & Real-life tasks - no task assignment & 12 males with ages ranging from 20 to 45 years old
 & 55 days & keyboard, mouse, application statistics and resource consumption & \xmark & \xmark & \xmark & \cmark\\
\hline

Meyer et al.\cite{meyer2020detecting} & Detecting developers’ task switches and types & 2020 & 2 studies: interaction observations and self-reports & 25 professional software developers (12 for 1st and 13 for 2nd study) & Logs for total of 51.7 hours of work (4 hours per participant) & Extracted a total of 109 temporal and semantic features from the computer interaction Data & \xmark & \xmark & \cmark & \xmark\\
\hline

Pellegrin et al.\cite{pellegrin2021task} & Task estimation for software company employees & 2021 & Real-life tasks, mainly including programming, test, documentation, administration, and leisure & 18 employees of a Japanese software development company, 13 developers 5 team leaders & 2 months, 780044 actions & Active apps, active windows, window titles, right and left clicks, keystrokes & \xmark & \xmark & \xmark & Excerpt of data (Full data on reason- able request)\\
\hline

RLKWiC & Knowledge work support and PIM & 2023 & Real-life tasks with no restrictions & 8 RPTU students, 6 males and 2 females & 2 months, over 61000 events within 56 contexts & Explicated contexts, browsed webs, window titles, active apps, clipboard actions & \cmark & \cmark & \cmark & \cmark\\
\hline

\end{tabular}
\label{tab1}
\end{center}
\label{comparison table}
\end{table*}

\section{Data Collection} \label{DC}
In this section, the process of data collection is elucidated in more detail. First, the experiment design is explained along with general demographic information about the participants. Then, the two tools used to collect the participants' digital activities are introduced, followed by clarifying the procedure of the experiment through three phases.

\subsection{Experiment Design}
As mentioned in the previous sections, one of the main goals of this data collection is to achieve a dataset that represents the challenging aspects of real knowledge work. To this end, instead of capturing the data in an controlled setting, the experiment has been designed in a way that the data could be collected during real-life digital tasks of participants without restricting their activities to a set of limited predefined tasks.

The participants have been chosen from volunteer Iranian students of the University of Kaiserslautern-Landau (RPTU), including six males and two females aged between 23 to 35 years old, studying different programs ranging from bachelor to post-doctorate in varied subjects (e.g., computer science, mechanics, mathematics, cognitive science, etc.) either in English or in German. Note that although a considerable amount of captured activities are related to the participant's studies, they include a wider variety of real-life activities like searching for a job, planning a trip, and following a particular subject of interest as entertainment (e.g., cryptocurrency trading, soccer matches, etc.). 

\subsection{Tools} \label{tools}

The collection of the RLKWiC dataset was done using a tailored version of the cSpaces tool\cite{jilek2018context}, integrated with User Activity Tracker (UAT)\cite{UATdemo} for additional features.

cSpaces, short for Context Spaces, is a recently proposed solution as a cornerstone of the semantic desktop\cite{sauermann2005overview, sauermann2009gnowsis} by Jilek et al.\cite{jilek2018context}. Rooted in the PIMO, cSpaces aims to represent the user's mental model as accurately as possible by allowing for the interlinking of information items, such as files, emails, and bookmarks that might be related to each other in the user's mind but are separated on their computer. Every activity performed is associated with one or more contexts, and each information item stored on a device can be linked with one or more of these contexts. The system is designed to be intuitive, allowing users to create contexts and add or remove elements. This approach tries to offer a more organized way to manage information by providing a dynamic reorganization capability to reduce the cognitive load.

Since not all the experimental features of cSpaces were required for the purpose of this data collection, a dedicated version of cSpaces with only the essential features has been used for RLKWiC. cSpaces provided the participants with a sidebar through which they could easily manage and organize their information (visit the demo page for more information\cite{cSpacesDemo}), while all their activities within the sidebar coupled with other external activities (including clipboard events and window titles using UAT and browsed web pages using cSpaces browser companion) were being captured in the background. Using the cSpaces sidebar during the experiment, participants were able to create their own contexts, annotate them with personal tags, and associate their local files and web pages with context spaces. They were then able to easily switch among their contexts and access the contained items inside whenever they wished. During the data collection, participants could choose to actively use the cSpaces app (for example by adding information items to the contexts) or just use it to express their current context.



\subsection{Procedure}
In what follows, the whole procedure of the experiment will be elucidated within three main phases.\\

\subsubsection{Preparation phase}
Before starting the data collection, a set of volunteer students was taken into consideration. Then, to investigate how suitable they are for the experiment, a survey was conducted to collect initial information about the volunteers like how many hours they spend in front of their computer on a daily basis, along with some technical questions like what operating system they use and what is their default web browser. Finally, eight of the initial volunteers were selected to participate in the study. After installing the tracking tools on their systems, the participants were instructed in a tutorial session on how to correctly use the tools and integrate cSpaces into their daily work to organize their information. The participants then started to use the tools for a test round in which the collected data was observed by the experimenter from time to time to prevent wrong usage of the tools and answer participants' questions in case of any ambiguity.\\

\subsubsection{Collection Phase}

After installing the required tools and instructing the participants through the process, the main data collection was started. While the participants were using cSpaces through their daily digital tasks for around two months (from the end of May till the first week of July 2023), their data in explicated contexts was collected. During the experiment, the participants were asked to keep the tracking tools on and use the context organizer app regularly. They had the option to easily turn off the trackers whenever they did not want their activities to be tracked.\\

\subsubsection{Post collection Phase} \label{pc phase}
After the data collection, two further steps, namely post-filtering and session-annotation, were taken to achieve the desired dataset. \\

Post-filtering: Although the participants were able to easily stop the tracking whenever they did not want their data to be captured during the collection phase, they were also provided with the possibility to remove tracked information for which they forgot to temporarily disable tracking, or just in case of any possible concerns regarding confidential or personal information being published. For this purpose, a dedicated application was developed for post-filtering through which the participants could see an overview of all the collected data (including files, web pages, tags, clipboard events, and window titles) and indicate the items that they did not want to be included in the dataset. All the collected information from the selected items was then obfuscated in the dataset (by being replaced with the string "$@|@|@$"). Nevertheless, the corresponding tracked events were not completely removed from the dataset as the metadata itself could be informative for analyzing user behaviors. For instance, in case a participant selects a web page to be removed from the dataset, all the information including the URL and the content of that web page is obfuscated. Yet, the fact that this participant has viewed a web page at a specific time is available in the dataset. \\

Session-annotation: In order to make the dataset more valuable for future user studies, especially in the user activity and task recognition domain (e.g.\cite{koldijk2012real, vuong2017watching}), the participants were asked to provide some additional information to the collected data in the post-collection phase. For this purpose, all the collected events were grouped into contextual knowledge work sessions. Each time that a user had selected a context and worked in it for a proper amount of time has been considered as a session. A new session is created either if the user switches the context or if no event is captured for more than half an hour. The sessions with too few events were ignored as they could not be associated with meaningful Knowledge Work Actions (KWA). They mostly belong to the wrong context switches as the participants tended to switch from context A to B, but chose C by mistake and then switched to the correct one (B) right away. The set of selected sessions were then shown to the participant (via a developed session annotation app) and they were asked to associate one or more KWA to each (e.g., learning, acquisition, information search, etc.) based on the typology proposed by Reinhardt et al.\cite{reinhardt2011knowledge}. In the session annotation process, the participants were asked to first select a set of general KWA that represent their main activities during each context created in the collection phase. The selected general KWA of the context were assigned to each session within that context by default. However, the participants were provided with the possibility to assign different KWA to sessions, in case the KWA within a session did not completely match the general KWA of the corresponding context. They were also asked to state whether their activities within each session were related to the context or should be considered as noise (due to distraction from the current context. e.g. reading an incoming irrelevant email). Fig. \ref{app} shows a sample screenshot from the test round as an overview of the post-annotation app through which the participants have indicated the distractions (by specifying whether a session is in or out of context), and enriched the collected data with DBpedia (https://www.dbpedia.org) resources and KWA.

\begin{figure}[htbp]
\centerline{\includegraphics[width=\columnwidth]{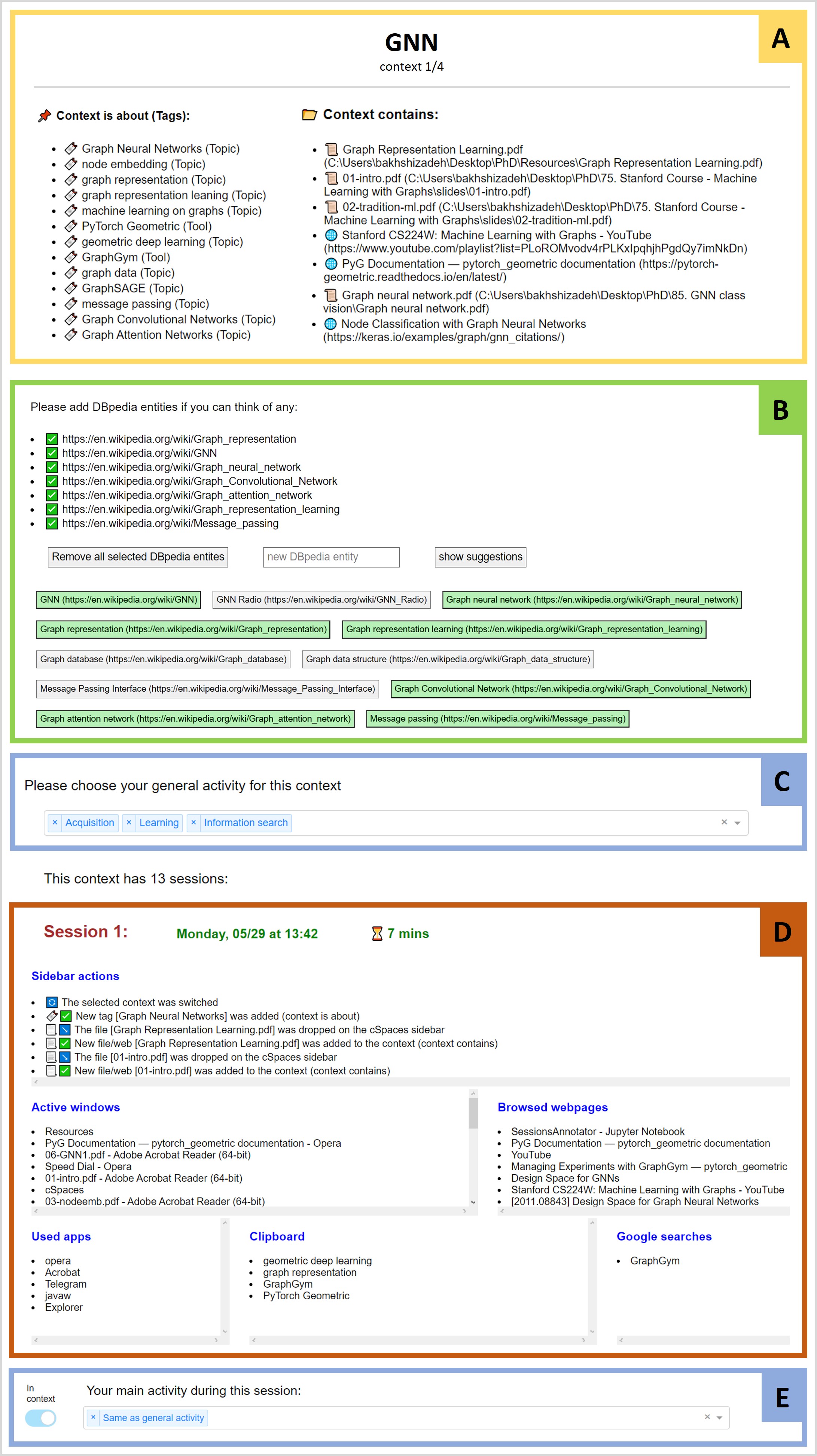}}
\caption{An overview of the post-annotation app consisted of five sections; A: Context information, B: Adding DBpedia resources to context with a suggestion feature, C: Selecting the general activity (set of KWA) for the context, D: An overview of a session in this sample context including various information, E: Session relevance (in/out of context) and session's activity.}
\label{app}
\end{figure}

In the end, the collected data was partially obfuscated based on the post-filtering, enriched with the post-annotation, and finally integrated into a structured dataset consisting of several features which will be discussed in the following section.

\section{Statistical overview and dataset structure} \label{DS}

During the approximately two months of the conducted study with eight volunteered participants involved, more than 61000 desktop events associated with 56 contexts (3-11 per participant) were collected. These explicated contexts were annotated with 211 personal concepts and 393 DBpedia resources, and contain 110 local files and more than 4200 browsed web pages. The built personal KGs in the background contained in total about 6400 involved nodes with more than 3100 inter-relations. The tracked desktop events were grouped into over 650 contextual sessions explicitly annotated with twelve different KWA.

The RLKWiC dataset includes various features categorized into five main components. In the following lines, the structure of the dataset will be elucidated through these five components:

\subsection{Contexts}
Contexts are the core component of RLKWiC around which the other components are established. The contexts are explicitly defined by the participants via the cSpaces sidebar. For each created context during the study, the participants assigned a label (i.e., name of the context), annotated them with their self-defined concepts as tags (plus DBpedia resources in the post-collection phase), and added their local files and web pages. The contexts are enriched with further features like the label's term vector (consisting of single-word terms), the number of events and sessions associated with that context, the creation and last selected timestamp, and assigned KWA. A sample context from RLKWiC associated with a university course that the corresponding participant was taking during the data collection is shown in Table \ref{sample context}.

\begin{table}[htbp]
\caption{a sample context from the RLKWiC Dataset}
\begin{center}
\begin{tabular}{p{2cm}|p{5.2cm}}
\hline
\textbf{Feature} & \textbf{Value} \\
\hline
context{\_}id & 12 \\
kg{\_}resid & 493 \\
label & Vehicle Dynamics Control \\
termvec{\_}label & (1050, 1), (2069, 1), (2137, 1) \\
stemtermvec{\_}label & (962, 1), (1882, 1), (1934, 1) \\
created & 1685280113957 \\
last{\_}selected & 1688635381032 \\
tags$^{\mathrm{a}}$ & Control (Topic), MEC Kaiserslautern (Group) \\
kg{\_}resources & 494, 495, 498 \\
files & 507 \\
webs & 408, 553 \\
KW{\_}action & Learning, Acquisition \\
{\#}events & 1240 \\
{\#}sessions & 8 \\
DBpedia$^{\mathrm{b}}$ & Control, Vehicle{\_}Dynamics{\_}Control \\
\multicolumn{2}{l}{$^{\mathrm{a}}$Tags are stored with IDs, here are shown with their labels and types.} \\
\multicolumn{2}{l}{$^{\mathrm{b}}$DBpedia entities are stored with URIs, here shown with their names.} \\
\end{tabular}
\label{sample context}
\end{center}
\end{table}

\subsection{Files and Web pages}

The documents in RLKWiC are either local files that participants have explicitly added to their contexts or the web pages they browsed during the data collection. Each web page is also associated with a context. Either explicitly, when added to a context by the participant, or implicitly, when browsed during a selected context. In the latter case, there is a possibility of irrelevant documents being included, which is one of the main challenges with real-life knowledge work. Files and web pages are stored as separate tables per participant. The features representing them include ID, path and name (for files), title and URL (for web pages), creation timestamp, and the corresponding term and concept vectors. The contents of the documents (in case they were not obfuscated for privacy reasons by participants) are stored separately as text files included in the dataset.

\subsection{Terms and Semantics}

RLKWiC is enriched with term and concept vectors coupled with personal KGs. Terms are 1grams extracted from textual contents included in the dataset and stored as terms and stemmed terms table (English and German stop words excluded). Concepts are created explicitly by the participants as tags for contexts, plus a few initial concepts in cSpaces.

Each document in the dataset has two vector representations. A concept vector indicates which of these concepts have appeared in the document and a term vector consists of the appeared terms. Each vector is shaped as an array of tuples representing the id of the appeared term or concept plus the frequency of its appearance in that document.

The built personal KGs in the background consist of defined contexts and concepts, added documents, and the relationships among them. These KGs are stored in RLKWiC as two tables. First, KG{\_}Resources which is the list of KG nodes represented by ID, URI, creation timestamp, label, and type ID. Second, subject-predicate-object triples (KG{\_}SPOs) that represent the relationships among the resources including the creation timestamp. Every item in the dataset is identified with a native ID, and a KG ID in case a node in the KG is associated with that resource. All created contexts and defined concepts are associated with a KG ID. Not all browsed documents included in the dataset are present in the KG, but only the ones that were explicitly added to the contexts by participants.

\subsection{Events}

All the captured activities of the participants during the data collection have been stored as events. Events can be considered as the basic units of RLKWiC and are captured by either UAT or cSpaces (introduced in \ref{tools}). Clipboard events and window titles are reported by UAT, activities inside the cSpaces sidebar (e.g. creating a new context, adding tags, opening a file from the context) by the cSpaces sidebar app, and the browsed web pages are reported by cSpaces browser companion.

Each event in RLKWiC is represented by ID, received and processed timestamp, reporter, cause, title, address, selected context ID, active app, and an SPO ID in case the event has triggered the creation of such a relation (SPO) in the personal KG in the background.

\subsection{Sessions}

As explained in session-annotation in the post-collection phase (as explained in Section \ref{pc phase}), sessions are meaningful units of user activities including a group of events within a context. The sessions in RLKWiC are represented by several features such as the selected context, session number, and list of included events. They are also post-annotated with matching KWA and a binary parameter showing either if it was relevant or a distraction from the context. A sample session (with minor modifications) from RLKWiC is presented in Table \ref{sample session}.

\begin{table}[htbp]
\caption{a sample session from the RLKWiC Dataset }
\begin{center}
\begin{tabular}{p{2.4cm}|p{5.5cm}}
\hline
\textbf{Feature} & \textbf{Value} \\
\hline
session{\_}id & p1-c4-s21 \\
context{\_}id & 4 \\
session{\_}number & 21 (= 21st session in context 4 of participant 1) \\
events{\_}list  & 20230606075629523, ... , 20230606080116515 \\
in context & True \\
KW{\_}actions & Information search, Acquisition \\
start & Tuesday - 2023/06/06 - 08:01:16 \\
approximate{\_}duration & 1 hour and 15 mins \\
google{\_}searches & install geopandas anaconda \\
clipboard & conda install --channel conda-forge geopandas, conda install geopandas \\
sidebar{\_}actions & new item was added to the context [488: Setup PyCharm and GDAL for Image Analysis - YouTube], new item was added to the context [489: Installation GeoPandas documentation], the selected context was switched \\
window{\_}titles & Installation GeoPandas documentation, Setup PyCharm and GDAL for Image Analysis - YouTube \\
browsed{\_}web pages & https://geopandas.org/en/stable/ getting-started/install.html, https://www.youtube.com/ watch?v=C0U15fADyZw \\
used{\_}apps & SearchApp, cSpaces, javaw, pycharm64, chrome \\
\end{tabular}
\label{sample session}
\end{center}
\end{table}

\section{Conclusion and Future Work} \label{CFW}

RLKWiC is a dataset of real-life knowledge work in context which is collected by monitoring computer interactions of eight volunteers during a period of two months.

Being the first publicly available dataset providing several information dimensions, RLKWiC tries to help the community by first, setting a common ground for evaluating and benchmarking PIM services, and second, providing the required data for leveraging machine learning approaches in this domain.

RLKWiC is structured over five components namely contexts, documents (files and web pages), terms and semantics, events, and sessions. It features explicated user contexts and contains all mentioned information items' content (websites, files, etc.) coupled with additional semantic information. RLKWiC can be beneficial to various research lanes such as user behavior modeling, PID, context-based information retrieval, and developing PIM services and PKA to support KW.

Since it has been tried to capture the characteristics of real-life knowledge work, having multilingual textual content along with some noise and obfuscations included in RLKWiC was hard to avoid. The obfuscation was due to privacy concerns, and the noises could have derived from either missed context switches or distractions and performing mini tasks. However, these are some of the main challenges that researchers have to deal with towards developing practical solutions for knowledge work support.

In conclusion, RLKWiC may serve as a helpful dataset for the community, propelling advancements in the PIM domain. By offering an intricate representation of real-life knowledge work, it is anticipated that researchers and developers will be better equipped to address current challenges and innovate new solutions. Researchers and interested parties can access and utilize the RLKWiC dataset at https://purl.org/RLKWiC.

\section{Acknowledgements}
We extend our gratitude to the volunteers who participated in this study. Their willingness to contribute has been pivotal to the success of this research.

This work was funded by the BMBF project SensAI (grant no. 01IW20007) and the DFG project Managed Forgetting (project no. 318396700).

\bibliographystyle{IEEEtran}
\bibliography{references}

\begin{thebibliography}{10}
\providecommand{\url}[1]{#1}
\csname url@samestyle\endcsname
\providecommand{\newblock}{\relax}
\providecommand{\bibinfo}[2]{#2}
\providecommand{\BIBentrySTDinterwordspacing}{\spaceskip=0pt\relax}
\providecommand{\BIBentryALTinterwordstretchfactor}{4}
\providecommand{\BIBentryALTinterwordspacing}{\spaceskip=\fontdimen2\font plus
\BIBentryALTinterwordstretchfactor\fontdimen3\font minus \fontdimen4\font\relax}
\providecommand{\BIBforeignlanguage}[2]{{%
\expandafter\ifx\csname l@#1\endcsname\relax
\typeout{** WARNING: IEEEtran.bst: No hyphenation pattern has been}%
\typeout{** loaded for the language `#1'. Using the pattern for}%
\typeout{** the default language instead.}%
\else
\language=\csname l@#1\endcsname
\fi
#2}}
\providecommand{\BIBdecl}{\relax}
\BIBdecl

\bibitem{drucker1999knowledge}
P.~F. Drucker, ``Knowledge-worker productivity: The biggest challenge,'' \emph{California management review}, vol.~41, no.~2, pp. 79--94, 1999.

\bibitem{davenport2005thinking}
T.~H. Davenport, \emph{Thinking for a living: how to get better performances and results from knowledge workers}.\hskip 1em plus 0.5em minus 0.4em\relax Harvard Business Press, 2005.

\bibitem{reinhardt2011knowledge}
W.~Reinhardt, B.~Schmidt, P.~Sloep, and H.~Drachsler, ``Knowledge worker roles and actions—results of two empirical studies,'' \emph{Knowledge and process management}, vol.~18, no.~3, pp. 150--174, 2011.

\bibitem{soto2021observing}
M.~Soto, C.~Satterfield, T.~Fritz, G.~C. Murphy, D.~C. Shepherd, and N.~Kraft, ``Observing and predicting knowledge worker stress, focus and awakeness in the wild,'' \emph{International Journal of Human-Computer Studies}, vol. 146, p. 102560, 2021.

\bibitem{koldijk2016context}
S.~J. Koldijk, ``Context-aware support for stress self-management: from theory to practice,'' Ph.D. dissertation, [Sl: sn], 2016.

\bibitem{HolzMausBernardiRostanin05b}
H.~Holz, H.~Maus, A.~Bernardi, and O.~Rostanin, ``From lightweight, proactive information delivery to business process-oriented knowledge management,'' \emph{J. of Universal Knowledge Management}, vol.~2, no. 2005, pp. 101--127, 2005.

\bibitem{vuong2017watching}
T.~Vuong, G.~Jacucci, and T.~Ruotsalo, ``Watching inside the screen: Digital activity monitoring for task recognition and proactive information retrieval,'' \emph{Proceedings of the ACM on Interactive, Mobile, Wearable and Ubiquitous Technologies}, vol.~1, no.~3, pp. 1--23, 2017.

\bibitem{sappelli2017evaluation}
M.~Sappelli, S.~Verberne, and W.~Kraaij, ``Evaluation of context-aware recommendation systems for information re-finding,'' \emph{J. of the Association for Information Science and Tech}, vol.~68, no.~4, 2017.

\bibitem{vuong2022behavioral}
T.~Vuong, ``Behavioral task modeling for entity recommendation,'' \emph{Series of publications A/Department of Computer Science, University of Helsinki.}, 2022.

\bibitem{jacucci2021entity}
G.~Jacucci, P.~Daee, T.~Vuong, S.~Andolina, K.~Klouche, M.~Sj{\"o}berg, T.~Ruotsalo, and S.~Kaski, ``Entity recommendation for everyday digital tasks,'' \emph{ACM Transactions on Computer-Human Interaction (TOCHI)}, vol.~28, no.~5, pp. 1--41, 2021.

\bibitem{bakhshizadeh2021leveraging}
M.~Bakhshizadeh, C.~Jilek, H.~Maus, and A.~Dengel, ``Leveraging context-aware recommender systems for improving personal knowledge assistants by introducing contextual states.'' in \emph{LWDA}, 2021, pp. 1--12.

\bibitem{chernov2007converting}
S.~Chernov, E.~Minack, and P.~Serdyukov, ``Converting desktop into a personal activity dataset,'' in \emph{Proceedings of 9th Russian national research conference on digital libraries}, 2007, pp. 280--283.

\bibitem{dhar2013data}
V.~Dhar, ``Data science and prediction,'' \emph{Communications of the ACM}, vol.~56, no.~12, pp. 64--73, 2013.

\bibitem{gonzalez2004constant}
V.~M. Gonz{\'a}lez and G.~Mark, ``"constant, constant, multi-tasking craziness" managing multiple working spheres,'' in \emph{SIGCHI Conf. on Human factors in computing systems, Proc.}, 2004.

\bibitem{koldijk2014swell}
S.~Koldijk, M.~Sappelli, S.~Verberne, M.~A. Neerincx, and W.~Kraaij, ``The swell knowledge work dataset for stress and user modeling research,'' in \emph{Proceedings of the 16th international conference on multimodal interaction}, 2014, pp. 291--298.

\bibitem{sappelli2014collecting}
M.~Sappelli, S.~Verberne, S.~Koldijk, and W.~Kraaij, ``Collecting a dataset of information behaviour in context,'' in \emph{4th Workshop on Context-Awareness in Retrieval and REC, Proc.}, 2014.

\bibitem{mirza2015study}
H.~T. Mirza, L.~Chen, I.~Hussain, A.~Majid, and G.~Chen, ``A study on automatic classification of users’ desktop interactions,'' \emph{Cybernetics and Systems}, vol.~46, no.~5, pp. 320--341, 2015.

\bibitem{satterfield2020identifying}
C.~Satterfield, T.~Fritz, and G.~C. Murphy, ``Identifying and describing information seeking tasks,'' in \emph{Proceedings of the 35th IEEE/ACM International Conference on Automated Software Engineering}, 2020, pp. 797--808.

\bibitem{sanchez2020behacom}
P.~M.~S. S{\'a}nchez, J.~M.~J. Valero, M.~Zago, A.~H. Celdr{\'a}n, L.~F. Maim{\'o}, E.~L. Bernal, S.~L. Bernal, J.~M. Valverde, P.~Nespoli, J.~P. Galindo \emph{et~al.}, ``Behacom-a dataset modelling users’ behaviour in computers,'' \emph{Data in Brief}, vol.~31, p. 105767, 2020.

\bibitem{pellegrin2021task}
F.~Pellegrin, Z.~Y{\"u}cel, A.~Monden, and P.~Leelaprute, ``Task estimation for software company employees based on computer interaction logs,'' \emph{Empirical Software Engineering}, vol.~26, pp. 1--48, 2021.

\bibitem{meyer2020detecting}
A.~N. Meyer, C.~Satterfield, M.~Z{\"u}ger, K.~Kevic, G.~C. Murphy, T.~Zimmermann, and T.~Fritz, ``Detecting developers’ task switches and types,'' \emph{IEEE Transactions on Software Engineering}, vol.~48, no.~1, pp. 225--240, 2020.

\bibitem{sauermann2007pimo}
L.~Sauermann, L.~Van~Elst, and A.~Dengel, ``Pimo-a framework for representing personal information models,'' \emph{Proceedings of I-Semantics}, vol.~7, pp. 270--277, 2007.

\bibitem{jilek2023towards}
C.~Jilek, M.~Schr{\"o}der, H.~Maus, S.~Schwarz, and A.~Dengel, ``Towards self-organizing personal knowledge assistants in evolving corporate memories,'' \emph{arXiv preprint arXiv:2308.01732}, 2023.

\bibitem{jilek2018context}
C.~Jilek, M.~Schr{\"o}der, S.~Schwarz, H.~Maus, and A.~Dengel, ``Context spaces as the cornerstone of a near-transparent and self-reorganizing semantic desktop,'' in \emph{The Semantic Web: ESWC 2018 Satellite Events: ESWC 2018 Satellite Events, Heraklion, Crete, Greece, June 3-7, 2018, Revised Selected Papers 15}.\hskip 1em plus 0.5em minus 0.4em\relax Springer, 2018, pp. 89--94.

\bibitem{UATdemo}
\BIBentryALTinterwordspacing
M.~Schr{\"o}der, ``User activity tracking,'' dfki.de, 2022. [Online]. Available: \url{https://www.dfki.uni-kl.de/kwt/user-activity-tracking/}
\BIBentrySTDinterwordspacing

\bibitem{sauermann2005overview}
L.~Sauermann, A.~Bernardi, and A.~Dengel, ``Overview and outlook on the semantic desktop,'' in \emph{Semantic Desktop Workshop; ISWC 2005, Proc.}, vol. 175.\hskip 1em plus 0.5em minus 0.4em\relax CEUR-WS.org, 2005.

\bibitem{sauermann2009gnowsis}
L.~Sauermann and M.~Jazayeri, ``The gnowsis semantic desktop approach to personal information management,'' Ph.D. dissertation, Department of Computer Science at University of Kaiserslautern-Landau (RPTU), 2009.

\bibitem{cSpacesDemo}
\BIBentryALTinterwordspacing
C.~Jilek, M.~Schr{\"o}der, S.~Schwarz, H.~Maus, and A.~Dengel, ``Context spaces as the cornerstone of a near-transparent and self-reorganizing semantic desktop,'' dfki.de, 2019. [Online]. Available: \url{https://www.dfki.uni-kl.de/~jilek/demo/cspaces/}
\BIBentrySTDinterwordspacing

\bibitem{koldijk2012real}
S.~Koldijk, M.~Van~Staalduinen, M.~Neerincx, and W.~Kraaij, ``Real-time task recognition based on knowledge workers' computer activities,'' in \emph{Proceedings of the 30th European Conference on Cognitive Ergonomics}, 2012, pp. 152--159.

\end{thebibliography}

\end{document}